\documentclass{article}

\usepackage{microtype}
\usepackage{graphicx}
\usepackage{subfigure}
\usepackage{booktabs}
\usepackage{multirow}
\usepackage{hyperref}
\usepackage{amsmath}
\usepackage{amssymb}
\usepackage{mathtools}
\usepackage{array}
\usepackage{enumitem}
\usepackage[capitalize, noabbrev]{cleveref}
\usepackage{float}
\usepackage{wrapfig}
\usepackage{booktabs} 
\usepackage{tabularx} 
\usepackage{adjustbox} 
\usepackage{makecell} 
\usepackage{multirow}
\usepackage{lipsum}
\usepackage{xcolor,soul}

\usepackage[accepted]{icml2026}
\makeatletter
\renewcommand{\ICML@appearing}{}
\makeatother

\icmltitlerunning{Efficient Multimodal Clinical Question Answering for Pulmonary Embolism Risk Assessment}

\begin{document}

\twocolumn[
\icmltitle{Efficient Multimodal Clinical Question Answering for Pulmonary Embolism Risk Assessment}

\begin{icmlauthorlist}
    \icmlauthor{Xiangyuan Xue}{aff1}
    \icmlauthor{Yang Yu}{aff1}
    \icmlauthor{Yan Gao}{cambridge}
    \icmlauthor{Junyan Wang}{adelaide}
    \icmlauthor{Bin Chen}{melbourne}
    \icmlauthor{Lingyan Ruan}{melbourne}
    \icmlauthor{Ting Dang}{melbourne}
    \icmlauthor{Hong Jia}{aff1}
\end{icmlauthorlist}

\icmlaffiliation{aff1}{University of Auckland, Auckland, New Zealand}
\icmlaffiliation{cambridge}{University of Cambridge, Cambridge, United Kingdom}
\icmlaffiliation{adelaide}{University of Adelaide, Adelaide, Australia}
\icmlaffiliation{melbourne}{University of Melbourne, Melbourne, Australia}

\icmlcorrespondingauthor{Hong Jia}{hong.jia@aucklanduni.ac.nz}

\icmlkeywords{Multimodal Large Language Models, Clinical Question Answering, Pulmonary Embolism, Electronic Health Records}

\vskip 0.3in
]

\printAffiliationsAndNotice{}

\begin{abstract}
Pulmonary embolism (PE) is a high risk cardiopulmonary condition whose management requires both timely diagnosis and reliable assessment of future clinical risk. Because PE care routinely combines computed tomography pulmonary angiography (CTPA), radiology interpretation, and longitudinal electronic health record (EHR) evidence, it provides a clinically meaningful setting for evaluating compact multimodal language models. In this work, we build a benchmark using efficient multimodal large language models (MLLMs) on INSPECT, a multimodal PE dataset containing 23,248 CTPA studies from 19,402 patients. We formulate eight diagnostic and prognostic tasks as structured clinical question answering problems and evaluate on typical efficient MLLMs under \texttt{CTPA-Only}, \texttt{\texttt{EHR-Only}}, and \texttt{\texttt{CTPA+EHR}} settings with zero-shot and few-shot prompting. Results show that Gemma4 E4B and Gemma4 E2B
perform more strongly when EHR evidence is available, especially under \texttt{\texttt{CTPA+EHR}} input. 
Task level analysis further shows that PE diagnosis achieves higher performance than prognostic tasks, particularly readmission prediction.
These observations suggest that compact multimodal models have the great potential in early stage PE risk detection and explanation.
\end{abstract}

\section{Introduction}
\label{sec:intro}

Pulmonary embolism (PE) is an urgent cardiopulmonary condition caused by obstruction of the pulmonary arterial circulation. It most often arises from thromboembolic material originating in the venous system. Its clinical burden reflects the combination of acute mortality risk, difficult early recognition, and possible complications after the index event. Patients may present with nonspecific symptoms, including dyspnea, chest pain, syncope, or otherwise unexplained deterioration. Diagnostic delay therefore remains a persistent challenge in routine care \citep{horlander2003pulmonary,alonso2010delay,hendriksen2017clinical}. Computed tomography pulmonary angiography (CTPA) is now a core imaging modality for suspected PE \citep{le2004diagnosing}. Yet CTPA interpretation in clinical practice is not based on imaging alone. Radiologists and physicians interpret imaging findings together with patients' diagnostic history, medication exposure, laboratory measurements, vital signs and demographic risk factors.
 These contextual factors can affect radiological interpretation and reporting, indicating that robust PE assessment is multimodal rather than purely visual \citep{leslie2000influence,cohen2007accuracy}.


Multimodal large language models (MLLMs) are a natural choice for evidence-based PE diagnosis and prognosis because they can align visual evidence from the index CTPA with patient-specific clinical information. In the diagnostic setting, the task is to determine whether PE is present on the index CTPA exam. 
In the prognostic setting, the task is to estimate the risk of later outcomes, including mortality, readmission, and pulmonary hypertension. Prognostic outcomes may depend on frailty, comorbidity burden, prior clinical instability, and treatment history, all of which are often captured more directly in EHRs than in the index CTPA exam.
 Although prior work has shown the value of combining CTPA and EHR information for PE-related prediction \citep{huang2020fusion,huang2020multimodal}, most existing approaches remain conventional score-based prediction systems rather than clinically grounded question-answering frameworks.

Beyond multimodal input integration, a question-answering formulation provides an additional way to structure PE assessment around explicit clinical questions and constrained answers.
Instead of defining separate task-specific classifier heads, this formulation represents each endpoint with a targeted clinical question, patient-specific evidence, and a constrained output schema. 
This is closer to clinical reasoning than independent risk scores, because the model must answer a specific clinical question based on the available patient evidence. The same question format can be applied to \texttt{CTPA-Only}, \texttt{EHR-Only}, and \texttt{CTPA+EHR} input settings. 
However, PE assessment remains challenging because the input may include volumetric CTPA images and long longitudinal EHR records.
This formulation also exposes evaluation challenges that may be hidden by standard classification metrics. In addition to this input complexity, the prediction endpoints are often imbalanced, with less frequent positive cases carrying substantial clinical importance. Under such imbalance, high accuracy may reflect repeated majority-class prediction rather than useful PE diagnosis or prognostic risk stratification. Therefore, evaluation should consider discrimination, positive-class recovery, modality-specific performance, and task-level variation, rather than relying only on average accuracy. 

In this work, we investigate multimodal PE diagnosis and prognosis prediction on INSPECT \citep{huang2023inspect}, a large-scale benchmark containing 23,248 CTPA studies from 19,402 patients. The dataset links imaging studies with radiology report impressions and structured longitudinal EHR records. INSPECT defines one diagnostic task, PE presence at the index CTPA, and seven prognostic tasks. The prognostic tasks cover in-hospital mortality, readmission, and pulmonary hypertension at clinically relevant time horizons. We reformulate the eight endpoints spanning PE diagnosis and prognosis as structured clinical question-answering tasks for compact MLLMs.
For each patient study, the MLLM
receives a task-specific question and evidence under one of three modality settings: \texttt{CTPA-Only}, \texttt{EHR-Only}, or \texttt{CTPA+EHR}. Each MLLM returns a binary prediction and a confidence score. We evaluate zero-shot and four-shot prompting across Qwen3.5 4B, Gemma4 E4B, Gemma4 E2B, and MedGemma using AUROC, AUPRC, accuracy, and F1.


Experiments show that compact MLLMs used for clinical question answering can support PE-related risk prediction when provided with appropriate clinical evidence.
The best performance is obtained when EHR information is available, and combining CTPA and EHR evidence yields the best zero-shot performance, with Gemma4 E4B achieving the highest AUROC. Four-shot prompting further improves positive-case recovery for the strongest Gemma configurations. Endpoint-level analysis shows that PE diagnosis is more effectively modeled than longitudinal outcomes such as readmission, suggesting that diagnostic endpoints are more directly supported by the available imaging and clinical context, whereas prognosis may require richer longitudinal information. Overall, these findings support the use of efficient MLLMs for PE-related clinical question answering while highlighting the importance of matching each clinical endpoint with the most informative evidence source.

The three contributions of this study are summarized below:
\begin{itemize}
    \item We provide a compact MLLM benchmark for multimodal PE diagnosis and prognosis in INSPECT, reformulating eight diagnostic and prognostic endpoints as structured clinical question-answering tasks across \texttt{CTPA-Only}, \texttt{EHR-Only}, and \texttt{CTPA+EHR} settings.
    \item We evaluate zero-shot and four-shot prompting across four compact MLLMs, showing that in-context examples improve performance only when the MLLMs can already interpret the relevant clinical evidence.
    \item We identify model and evaluation failure modes in compact multimodal medical QA, including majority-class collapse, misleadingly high accuracy under class imbalance, weak \texttt{CTPA-Only} performance, and task-level heterogeneity between PE diagnosis and longitudinal prognosis.

\end{itemize}

\section{Related Work}
\label{sec:related}

\textbf{Deep Learning for PE Risk Prediction.}
Early deep learning studies in pulmonary embolism (PE) primarily focused on automated diagnosis from computed tomography pulmonary angiography (CTPA) scans \citep{soffer2021deep, liu2020evaluation, huang2020penet}. In particular, end-to-end models based on 3D convolutional neural networks (CNNs) have achieved strong performance in detecting emboli within volumetric CTPA images \citep{huang2020penet, colak2021rsna}. More recently, multimodal learning approaches that integrate 3D CTPA imaging with longitudinal electronic health records (EHRs) have been introduced to leverage complementary clinical information for improved PE diagnosis and risk assessment \citep{huang2020fusion}. 
However, existing multimodal PE models are designed to produce task-specific probability scores or risk estimates, rather than to generate explanations, answer clinician questions, or support interactive reasoning over patient data. This limits their interpretability and usability in clinical decision-making.

\textbf{Multimodal Medical QA.}
Recent studies have explored Medical Visual Question Answering (Med-VQA) as a means of supporting clinical reasoning over medical images and associated patient information \citep{he2021towards, liu2021slake}. Progress has been driven largely by multimodal large language models (MLLMs)  \citep{moor2023med}, including LLaVA-Med \citep{li2023llava}, Med-Flamingo \citep{moor2023med}, and Med-PaLM M \citep{tu2024towards}, which have shown promising capabilities in medical image interpretation, clinical question answering, and report generation. By aligning visual representations from medical images with text-based clinical information, these models can generate natural-language responses to complex clinical queries. However, most state-of-the-art MLLMs rely on cloud-based inference, which may require transmitting sensitive patient data to external servers and therefore raises privacy concerns \citep{price2019privacy, sangaraju2025ai, clusmann2023future}. Although local deployment could reduce these data-sharing risks, large-scale MLLMs typically require substantial computational resources that are not available in many clinical environments. This limits their widespread deployment in resource-constrained healthcare settings. Closest to our setting, ~\citet{zhong2025ctpa} used CTPA-based MLLMs to predict fine-grained PE-related abnormalities for report generation and outcome prediction, while Uniferum~\citep{lee2025uniferum} evaluated a volumetric CT VLM on INSPECT classification tasks in a zero-shot setting. Our work is complementary: rather than proposing a new volumetric encoder or report-generation pipeline, we reformulate all eight INSPECT endpoints as compact-model clinical QA tasks and compare \texttt{CTPA-Only}, \texttt{EHR-Only}, and \texttt{CTPA+EHR }evidence.

\section{INSPECT Dataset}
\label{sec:inspect}

INSPECT, short for Integrating Numerous Sources for Prognostic Evaluation of Clinical Timelines, is a multimodal dataset for pulmonary embolism diagnosis and prognosis \citep{huang2023inspect}. It contains 23,248 CTPA studies from 19,402 adult patients and links each index CTPA study with radiology report impressions and structured longitudinal EHR records. Each case is anchored at the index CTPA, which serves as the reference point for both acute PE diagnosis and subsequent prognostic prediction. Core dataset statistics are summarized in Table~\ref{tab:inspect_stats}.

\begin{figure*}[t]
    \centering
    \includegraphics[width=\textwidth]{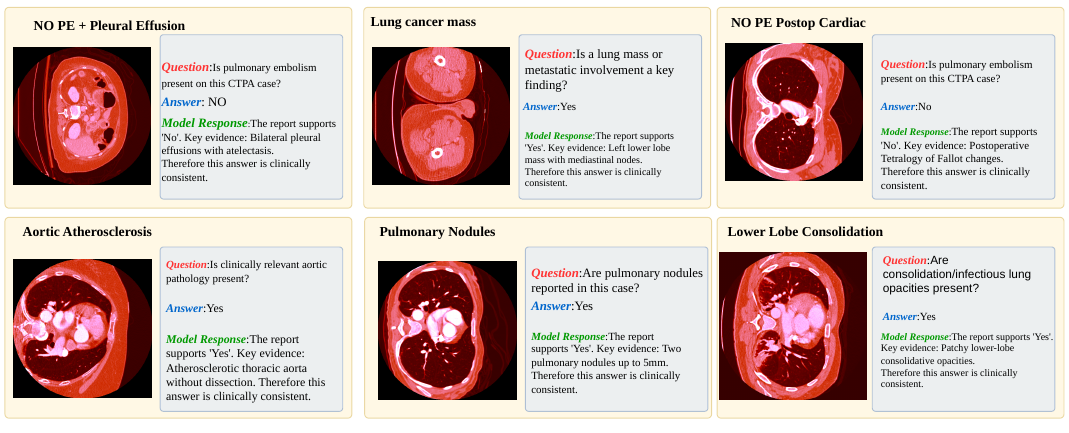}
    \caption{Representative structured clinical QA examples. Each panel shows the model-visible evidence type, task-specific question, reference answer for illustration only, and structured model response. Reference answers and radiology report impressions are not provided to the model during inference.}
    \label{fig:pe_qa_examples}
\end{figure*}

\subsection{Cohort and Modalities}
\label{sec:cohort_modalities}

The official INSPECT split is defined at the patient level, preventing the same patient from appearing in both training and evaluation partitions. The split contains 18,945 training studies, 1,089 validation studies, and 3,214 test studies, corresponding to 15,789 training patients, 913 validation patients, and 2,700 test patients. The dataset also accounts for overlap with earlier PE resources such as RSPECT and RadFusion to reduce the risk of evaluation contamination \citep{colak2021rsna,zhou2021radfusion}. 

INSPECT provides three complementary evidence sources. The first is volumetric CTPA imaging, which is the most direct modality for identifying embolic findings in the pulmonary arterial tree. The second is the radiology report impression, which is used in the original INSPECT construction for diagnostic label generation, including report derived PE labels based on a clinical language model validated against manually annotated radiology text \citep{banerjee2019comparative}. In our experiments, report impressions are described for dataset completeness but are not provided as direct model input, reducing the risk of label leakage. The third modality is structured longitudinal EHR data, including demographics, diagnoses, procedures, medications, laboratory measurements, vital signs, visits, and death records. This EHR component is especially relevant for prognosis, since mortality, readmission, and pulmonary hypertension after the index CTPA may depend on comorbidity burden and prior clinical trajectory rather than imaging findings alone. 

\begin{table}[t]
\centering
\small
\caption{Core INSPECT dataset statistics used in this study.}
\label{tab:inspect_stats}
\begin{tabular}{lr}
\toprule
\textbf{Attribute} & \textbf{Value} \\
\midrule
Patients & 19,402 \\
CTPA studies & 23,248 \\
Training studies & 18,945 \\
Validation studies & 1,089 \\
Test studies & 3,214 \\
Training patients & 15,789 \\
Validation patients & 913 \\
Test patients & 2,700 \\
Approximate CT slices & 5.16M \\
Structured EHR records & 225.44M \\
Input sources & CT, report, EHR \\
Benchmark tasks & 8 \\
\bottomrule
\end{tabular}
\end{table}
\subsection{Task Definitions}
\label{sec:tasks}

INSPECT defines one diagnostic task and seven prognostic tasks. The diagnostic task asks whether PE is present at the index CTPA. The prognostic tasks ask whether future clinical events occur after the index scan, including in hospital mortality at 1, 6, and 12 months, readmission at 1, 6, and 12 months, and pulmonary hypertension at 12 months. Prognostic samples without sufficient follow up are marked as censored and excluded from the corresponding task evaluation. The resulting label distributions, shown in Table~\ref{tab:task_stats}, are highly imbalanced, with positive outcomes substantially less frequent than negative outcomes across most tasks. This imbalance motivates our use of AUROC, AUPRC, and F1 in addition to accuracy.

\begin{table*}[t]
\centering
\small
\caption{INSPECT task definitions and cohort-level label counts. Censored samples are excluded from task-level evaluation.}
\label{tab:task_stats}
\begin{adjustbox}{max width=\textwidth}
\begin{tabular}{llrrr}
\toprule
\textbf{Task group} & \textbf{Endpoint} & \textbf{Positive} & \textbf{Negative} & \textbf{Censored} \\
\midrule
Diagnosis & Pulmonary embolism at index CTPA & 4,689 & 18,559 & 0 \\
Mortality & 1 month in-hospital mortality & 1,200 & 20,803 & 1,245 \\
Mortality & 6 month in-hospital mortality & 2,389 & 18,552 & 2,307 \\
Mortality & 12 month in-hospital mortality & 2,916 & 17,157 & 3,175 \\
Readmission & 1 month readmission & 857 & 20,774 & 1,617 \\
Readmission & 6 month readmission & 2,185 & 17,953 & 3,110 \\
Readmission & 12 month readmission & 2,826 & 16,253 & 4,169 \\
Pulmonary hypertension & 12 month pulmonary hypertension & 2,726 & 16,503 & 4,019 \\
\bottomrule
\end{tabular}
\end{adjustbox}
\end{table*}

\subsection{Clinical Question Answering Formulation}
\label{sec:qa_formulation}

The original INSPECT benchmark evaluates dedicated predictive models over image, EHR, and fused representations \citep{huang2023inspect}. In this work, we adapt the dataset to compact multimodal language models by reformulating each endpoint as a structured clinical question answering task. For each case, the model receives a task-specific question and modality-specific evidence under one of three settings: \texttt{CTPA-Only}, \texttt{EHR-Only}, or \texttt{\texttt{CTPA+EHR}}. The model is instructed to return a constrained structured response containing a binary answer and a confidence score. The binary answer is used for accuracy and F1, while the confidence score is used for AUROC and AUPRC. When a short rationale is generated, it is used only for qualitative inspection and is not used for metric computation. 

This formulation preserves the clinical meaning of the original endpoints while providing a unified interface for small multimodal models. It also makes failure modes easier to analyze. A model may fail by assigning uninformative confidence scores, by repeatedly predicting the majority class, by ignoring one modality, or by failing to connect the evidence to the requested time horizon. Figure~\ref{fig:pe_qa_examples} illustrates representative examples of this clinical question answering format.

\section{Experimental Setup}
\label{sec:experimental_setup}

\subsection{Data Split and Label Handling}
\label{sec:data_split_label}

All experiments are evaluated on the official INSPECT test split. The benchmark contains eight binary tasks: one diagnostic endpoint for PE presence at the index CTPA and seven prognostic endpoints covering 1, 6, and 12 month in hospital mortality, 1, 6, and 12 month readmission, and 12 month pulmonary hypertension. For prognostic tasks, samples marked as censored in INSPECT are excluded from the corresponding endpoint evaluation. We do not impute censored labels or treat them as negative cases. Metrics are first computed separately for each endpoint and then macro averaged across tasks, giving each clinical endpoint equal weight.
\vspace{-0.4cm}
\subsection{Models and Input Settings}
\label{sec:models_inputs}

We evaluate four compact multimodal language model checkpoints without parameter updates. Table~\ref{tab:model_config} reports the model identifiers and inference settings used in our experiments. All models are evaluated with deterministic decoding using \texttt{temperature=0}, \texttt{do\_sample=false}, and \texttt{max\_new\_tokens=128}. Each output is parsed into a binary prediction and a confidence score, which are used for classification and ranking based metrics, respectively.

\begin{table}[t]
\centering
\scriptsize
\setlength{\tabcolsep}{2.0pt}
\renewcommand{\arraystretch}{1.15}
\caption{
Model checkpoints and inference configuration used in the benchmark.
}
\label{tab:model_config}
\begin{adjustbox}{max width=\columnwidth}
\begin{tabular}{llll}
\toprule
\textbf{Model} & \textbf{Checkpoint ID} & \textbf{Scale} & \textbf{Precision} \\
\midrule
Qwen3.5 4B & \texttt{Qwen/Qwen3.5-4B} & 4B & bf16 \\
Gemma4 E4B & \texttt{google/gemma-4-e4b-it} & 4B & bf16 \\
Gemma4 E2B & \texttt{google/gemma-4-e2b-it} & 2B & bf16 \\
MedGemma & \texttt{google/medgemma-1.5-4b-it} & 4B & bf16 \\
\bottomrule
\end{tabular}
\end{adjustbox}
\end{table}

Each model is evaluated under three input settings. In the \texttt{CTPA-Only} setting, the model receives the CTPA-derived visual input and task question. Because the evaluated compact MLLMs are 2D-native, each volumetric CTPA study is converted into a deterministic 2D montage by selecting \(K\) axial slices using a fixed label independent rule, resizing them, and arranging them in anatomical order. In the \texttt{EHR-Only} setting, the model receives serialized structured EHR evidence and the task question, while a blank image placeholder is used only when required by the model interface. The EHR serialization includes only information available before the index CTPA timestamp and groups records into fixed sections such as demographics, diagnoses, procedures, medications, laboratory measurements, vital signs, and visit history. In the \texttt{CTPA+EHR} setting, the model receives both the CTPA montage and serialized EHR evidence. If an image is missing or fails preprocessing, we use the same blank placeholder rather than dropping the sample. Report impressions are withheld in all main experiments to reduce label leakage from report derived PE labels.

\subsection{Prompting Protocol}
\label{sec:prompting}

We compare zero-shot and four-shot prompting. In the zero-shot setting, the model receives only the instruction, task question, modality specific evidence, and required output schema. In the four-shot setting, four demonstrations are prepended before the test case. These demonstrations are fixed before evaluation, drawn only from the training split, and shared across models for the same task and modality setting. The demonstration order is also fixed. Patient level separation is enforced so that no patient used in a demonstration appears in the test split. This design keeps the few-shot setting reproducible and prevents leakage from test cases into the prompt.

\subsection{Evaluation Metrics and Completion Criteria}
\label{sec:metrics_completion}

We report AUROC, AUPRC, accuracy, and F1. AUROC measures threshold independent discrimination, while AUPRC is included because INSPECT endpoints are strongly imbalanced and positive cases are clinically important. Accuracy is reported for completeness, but it is not treated as the primary indicator of model utility because majority class prediction can produce deceptively high values. F1 is used to summarize positive class recovery under the model's discrete prediction.

A run is considered complete only when all eight INSPECT tasks are evaluated on the test split. The main ranking and all primary claims are based on complete runs only. Partial outputs are excluded from the main comparison and may be used only for diagnostic or supplementary analysis. All primary results are computed on the official INSPECT test split.

\vspace{-0.5cm}

\section{Results}
\label{sec:results}
\begin{table}[t]
\centering
\scriptsize
\setlength{\tabcolsep}{1.5pt}
\renewcommand{\arraystretch}{1.3}
\caption{
Zero-shot and four-shot results on the INSPECT test set. Each entry reports the macro average across the eight diagnostic and prognostic tasks, with $n_{\mathrm{sum}}=23{,}079$ for the completed task set. AUROC and AUPRC measure threshold-independent discrimination, while F1 reflects positive-class recovery under the model's discrete prediction. Accuracy is reported for completeness but should be interpreted cautiously because several INSPECT outcomes are highly imbalanced. Bold and underlined values denote the best and second-best completed runs, respectively, for each metric column.
}
\label{tab:main_results_zero_fewshot}
\begin{adjustbox}{max width=\textwidth}
\begin{tabular}{lrrrrrrrr}
\toprule
\multirow{2}{*}{\textbf{Model}} 
& \multicolumn{4}{c}{\textbf{Zero-shot}} 
& \multicolumn{4}{c}{\textbf{Four-shot}} \\
\cmidrule(lr){2-5}
\cmidrule(lr){6-9}
& \textbf{AUROC} & \textbf{AUPRC} & \textbf{Acc.} & \textbf{F1}
& \textbf{AUROC} & \textbf{AUPRC} & \textbf{Acc.} & \textbf{F1} \\
\midrule

\multicolumn{9}{l}{\textbf{\texttt{CTPA-Only}}} \\
Qwen3.5 4B 
& 0.5000 & 0.1155 & \underline{0.8845} & 0.0000
& 0.5000 & 0.1155 & \textbf{0.8845} & 0.0000 \\
Gemma4 E4B 
& 0.4995 & 0.1156 & 0.8695 & 0.0263
& 0.5030 & 0.1164 & 0.8042 & 0.0419 \\
Gemma4 E2B 
& 0.4969 & 0.1154 & 0.2133 & 0.1708
& 0.4974 & 0.1147 & 0.2413 & 0.1974 \\
MedGemma 
& 0.5000 & 0.1155 & \underline{0.8845} & 0.0000
& 0.5000 & 0.1155 & \textbf{0.8845} & 0.0000 \\

\midrule
\multicolumn{9}{l}{\textbf{EHR-Only}} \\
Qwen3.5 4B 
& 0.5000 & 0.1155 & \underline{0.8845} & 0.0000
& 0.5000 & 0.1155 & 0.8796 & 0.0083 \\
Gemma4 E4B 
& \underline{0.6615} & \underline{0.2494} & \textbf{0.8891} & 0.2646
& 0.6748 & 0.2670 & 0.7455 & 0.3300 \\
Gemma4 E2B 
& 0.5465 & 0.1532 & 0.3235 & 0.2822
& 0.6566 & 0.2343 & 0.7049 & 0.3190 \\
MedGemma 
& 0.5006 & 0.1168 & 0.8795 & 0.0075
& 0.5011 & 0.1168 & 0.8795 & 0.0080 \\

\midrule
\multicolumn{9}{l}{\textbf{\texttt{\texttt{CTPA+EHR}}}} \\
Qwen3.5 4B 
& 0.5000 & 0.1155 & \underline{0.8845} & 0.0000
& 0.5000 & 0.1155 & \textbf{0.8845} & 0.0000 \\
Gemma4 E4B 
& \textbf{0.6893} & \textbf{0.2638} & 0.8153 & \textbf{0.3286}
& \textbf{0.6867} & \underline{0.2680} & 0.7284 & \textbf{0.3359} \\
Gemma4 E2B 
& 0.6024 & 0.1793 & 0.5343 & \underline{0.2963}
& \underline{0.6836} & \textbf{0.2706} & 0.7703 & \underline{0.3336} \\
MedGemma 
& 0.4992 & 0.1158 & 0.8823 & 0.0014
& 0.4899 & 0.1158 & \underline{0.8823} & 0.0015 \\

\bottomrule
\end{tabular}
\end{adjustbox}
\end{table}

\begin{table}[t]
\centering
\scriptsize
\setlength{\tabcolsep}{1.5pt}
\renewcommand{\arraystretch}{1.3}
\caption{
Task-level results for PE diagnosis on the INSPECT test set. The task evaluates PE presence at the index CTPA. Each row reports zero-shot and four-shot performance under the same model and modality setting.
}
\label{tab:diagnosis_results}
\begin{adjustbox}{max width=\columnwidth}
\begin{tabular}{lcccccccc}
\toprule
\multirow{2}{*}{\textbf{Model}}
& \multicolumn{4}{c}{\textbf{Zero-shot}}
& \multicolumn{4}{c}{\textbf{Four-shot}} \\
\cmidrule(lr){2-5}
\cmidrule(lr){6-9}
& \textbf{AUROC} & \textbf{AUPRC} & \textbf{Acc.} & \textbf{F1}
& \textbf{AUROC} & \textbf{AUPRC} & \textbf{Acc.} & \textbf{F1} \\
\midrule

\multicolumn{9}{l}{\textbf{\texttt{CTPA-Only}}} \\
Qwen3.5 4B
& 0.5000 & 0.1795 & 0.8205 & 0.0000
& 0.5000 & 0.1795 & 0.8205 & 0.0000 \\
Gemma4 E4B
& 0.4959 & 0.1805 & 0.7004 & 0.2100
& 0.5018 & 0.1800 & 0.1857 & 0.3049 \\
Gemma4 E2B
& 0.5278 & 0.1899 & 0.8043 & 0.0654
& 0.5000 & 0.1795 & 0.1795 & 0.3044 \\
MedGemma
& 0.5000 & 0.1795 & 0.8205 & 0.0000
& 0.5000 & 0.1795 & 0.8205 & 0.0000 \\

\midrule
\multicolumn{9}{l}{\textbf{EHR-Only}} \\
Qwen3.5 4B
& 0.5000 & 0.1795 & 0.8205 & 0.0000
& 0.5000 & 0.1795 & 0.8205 & 0.0000 \\
Gemma4 E4B
& \underline{0.9580} & \underline{0.8973} & \underline{0.9701} & \underline{0.9137}
& \underline{0.9681} & \underline{0.9123} & \textbf{0.9704} & \textbf{0.9151} \\
Gemma4 E2B
& 0.7952 & 0.4664 & 0.9415 & 0.8479
& 0.9325 & 0.8445 & 0.9564 & 0.8807 \\
MedGemma
& 0.5090 & 0.1877 & 0.8220 & 0.0205
& 0.5000 & 0.1795 & 0.8205 & 0.0000 \\

\midrule
\multicolumn{9}{l}{\textbf{\texttt{\texttt{CTPA+EHR}}}} \\
Qwen3.5 4B
& 0.5000 & 0.1795 & 0.8205 & 0.0000
& 0.5000 & 0.1795 & 0.8205 & 0.0000 \\
Gemma4 E4B
& \textbf{0.9683} & \textbf{0.9084} & \textbf{0.9726} & \textbf{0.9216}
& \textbf{0.9706} & \textbf{0.9150} & \underline{0.9683} & \underline{0.9101} \\
Gemma4 E2B
& 0.8497 & 0.5598 & 0.9530 & 0.8732
& 0.9467 & 0.8660 & 0.9571 & 0.8821 \\
MedGemma
& 0.4924 & 0.1796 & 0.8027 & 0.0063
& 0.4924 & 0.1796 & 0.8027 & 0.0063 \\

\bottomrule
\end{tabular}
\end{adjustbox}
\end{table}

\begin{table}[t]
\centering
\scriptsize
\setlength{\tabcolsep}{1.5pt}
\renewcommand{\arraystretch}{1.3}
\caption{
Task-group results for mortality prediction on the INSPECT test set. Values are macro-averaged over 1-month, 6-month, and 12-month in-hospital mortality tasks.
}
\label{tab:mortality_results}
\begin{adjustbox}{max width=\columnwidth}
\begin{tabular}{lcccccccc}
\toprule
\multirow{2}{*}{\textbf{Model}}
& \multicolumn{4}{c}{\textbf{Zero-shot}}
& \multicolumn{4}{c}{\textbf{Four-shot}} \\
\cmidrule(lr){2-5}
\cmidrule(lr){6-9}
& \textbf{AUROC} & \textbf{AUPRC} & \textbf{Acc.} & \textbf{F1}
& \textbf{AUROC} & \textbf{AUPRC} & \textbf{Acc.} & \textbf{F1} \\
\midrule

\multicolumn{9}{l}{\textbf{\texttt{CTPA-Only}}} \\
Qwen3.5 4B
& 0.5000 & 0.1021 & \textbf{0.8979} & 0.0000
& 0.5000 & 0.1021 & \textbf{0.8979} & 0.0000 \\
Gemma4 E4B
& 0.5000 & 0.1021 & \textbf{0.8979} & 0.0000
& 0.5114 & 0.1043 & \underline{0.8958} & 0.0059 \\
Gemma4 E2B
& 0.5000 & 0.1021 & 0.1021 & 0.1832
& 0.4945 & 0.1014 & 0.1120 & 0.1815 \\
MedGemma
& 0.5000 & 0.1021 & \textbf{0.8979} & 0.0000
& 0.5000 & 0.1021 & \textbf{0.8979} & 0.0000 \\

\midrule
\multicolumn{9}{l}{\textbf{EHR-Only}} \\
Qwen3.5 4B
& 0.5000 & 0.1021 & \textbf{0.8979} & 0.0000
& 0.5000 & 0.1021 & 0.8848 & 0.0220 \\
Gemma4 E4B
& \underline{0.6855} & \underline{0.1721} & 0.8788 & \underline{0.2226}
& \underline{0.6995} & \underline{0.1935} & 0.7659 & \underline{0.2905} \\
Gemma4 E2B
& 0.5233 & 0.1064 & 0.1514 & 0.1897
& 0.6757 & 0.1612 & 0.6561 & 0.2666 \\
MedGemma
& 0.4997 & 0.1022 & \underline{0.8930} & 0.0055
& 0.5000 & 0.1021 & \textbf{0.8979} & 0.0000 \\

\midrule
\multicolumn{9}{l}{\textbf{\texttt{\texttt{CTPA+EHR}}}} \\
Qwen3.5 4B
& 0.5000 & 0.1021 & \textbf{0.8979} & 0.0000
& 0.5000 & 0.1021 & \textbf{0.8979} & 0.0000 \\
Gemma4 E4B
& \textbf{0.7242} & \textbf{0.1963} & 0.8055 & \textbf{0.3049}
& \textbf{0.7240} & \textbf{0.1995} & 0.7460 & \textbf{0.2966} \\
Gemma4 E2B
& 0.5866 & 0.1212 & 0.3216 & 0.2117
& 0.6789 & 0.1701 & 0.6867 & 0.2666 \\
MedGemma
& 0.5000 & 0.1021 & \textbf{0.8979} & 0.0000
& 0.5000 & 0.1021 & \textbf{0.8979} & 0.0000 \\

\bottomrule
\end{tabular}
\end{adjustbox}
\end{table}

\begin{table}[t]
\centering
\scriptsize
\setlength{\tabcolsep}{1.5pt}
\renewcommand{\arraystretch}{1.3}
\caption{
Task-group results for readmission prediction on the INSPECT test set. Values are macro-averaged over 1-month, 6-month, and 12-month readmission tasks. Compared with PE diagnosis and mortality prediction, readmission remains the most difficult group for compact models.
}
\label{tab:readmission_results}
\begin{adjustbox}{max width=\columnwidth}
\begin{tabular}{lcccccccc}
\toprule
\multirow{2}{*}{\textbf{Model}}
& \multicolumn{4}{c}{\textbf{Zero-shot}}
& \multicolumn{4}{c}{\textbf{Four-shot}} \\
\cmidrule(lr){2-5}
\cmidrule(lr){6-9}
& \textbf{AUROC} & \textbf{AUPRC} & \textbf{Acc.} & \textbf{F1}
& \textbf{AUROC} & \textbf{AUPRC} & \textbf{Acc.} & \textbf{F1} \\
\midrule

\multicolumn{9}{l}{\textbf{\texttt{CTPA-Only}}} \\
Qwen3.5 4B
& 0.5000 & 0.1012 & \underline{0.8988} & 0.0000
& 0.5000 & 0.1012 & \underline{0.8988} & 0.0000 \\
Gemma4 E4B
& 0.5000 & 0.1012 & \underline{0.8988} & 0.0000
& 0.4972 & 0.1014 & 0.8983 & 0.0043 \\
Gemma4 E2B
& 0.4981 & 0.1009 & 0.1296 & \underline{0.1798}
& 0.4984 & 0.0996 & 0.4265 & 0.1641 \\
MedGemma
& 0.5000 & 0.1012 & \underline{0.8988} & 0.0000
& 0.5000 & 0.1012 & \underline{0.8988} & 0.0000 \\

\midrule
\multicolumn{9}{l}{\textbf{EHR-Only}} \\
Qwen3.5 4B
& 0.5000 & 0.1012 & \underline{0.8988} & 0.0000
& 0.5000 & 0.1012 & \underline{0.8988} & 0.0000 \\
Gemma4 E4B
& \underline{0.5310} & 0.1045 & 0.8814 & 0.0735
& 0.5470 & 0.1170 & 0.6149 & 0.1744 \\
Gemma4 E2B
& 0.5091 & 0.1026 & 0.1744 & \textbf{0.1811}
& 0.5500 & 0.1130 & 0.6318 & \underline{0.1835} \\
MedGemma
& 0.4989 & 0.1016 & 0.8900 & 0.0077
& 0.4989 & 0.1016 & 0.8900 & 0.0077 \\

\midrule
\multicolumn{9}{l}{\textbf{\texttt{\texttt{CTPA+EHR}}}} \\
Qwen3.5 4B
& 0.5000 & 0.1012 & \underline{0.8988} & 0.0000
& 0.5000 & 0.1012 & \underline{0.8988} & 0.0000 \\
Gemma4 E4B
& \textbf{0.5593} & \textbf{0.1175} & 0.7622 & 0.1564
& \underline{0.5541} & \underline{0.1177} & 0.5931 & 0.1791 \\
Gemma4 E2B
& 0.5277 & \underline{0.1082} & 0.5155 & 0.1774
& \textbf{0.6141} & \textbf{0.1937} & 0.7754 & \textbf{0.2221} \\
MedGemma
& 0.5004 & 0.1018 & \textbf{0.8990} & 0.0016
& 0.5004 & 0.1018 & \textbf{0.8990} & 0.0016 \\

\bottomrule
\end{tabular}
\end{adjustbox}
\end{table}

\begin{table}[t]
\centering
\scriptsize
\setlength{\tabcolsep}{1.5pt}
\renewcommand{\arraystretch}{1.3}
\caption{
Task-level results for 12-month pulmonary hypertension prediction on the INSPECT test set. This endpoint evaluates longer-horizon cardiopulmonary risk after the index CTPA study.
}
\label{tab:ph_results}
\begin{adjustbox}{max width=\columnwidth}
\begin{tabular}{lcccccccc}
\toprule
\multirow{2}{*}{\textbf{Model}}
& \multicolumn{4}{c}{\textbf{Zero-shot}}
& \multicolumn{4}{c}{\textbf{Four-shot}} \\
\cmidrule(lr){2-5}
\cmidrule(lr){6-9}
& \textbf{AUROC} & \textbf{AUPRC} & \textbf{Acc.} & \textbf{F1}
& \textbf{AUROC} & \textbf{AUPRC} & \textbf{Acc.} & \textbf{F1} \\
\midrule

\multicolumn{9}{l}{\textbf{\texttt{CTPA-Only}}} \\
Qwen3.5 4B
& 0.5000 & 0.1348 & \textbf{0.8652} & 0.0000
& 0.5000 & 0.1348 & \textbf{0.8652} & 0.0000 \\
Gemma4 E4B
& 0.5000 & 0.1348 & \textbf{0.8652} & 0.0000
& 0.4963 & 0.1340 & \textbf{0.8652} & 0.0000 \\
Gemma4 E2B
& 0.4533 & 0.1246 & 0.2071 & 0.2121
& 0.5002 & 0.1349 & 0.1352 & 0.2377 \\
MedGemma
& 0.5000 & 0.1348 & \textbf{0.8652} & 0.0000
& 0.5000 & 0.1348 & \textbf{0.8652} & 0.0000 \\

\midrule
\multicolumn{9}{l}{\textbf{EHR-Only}} \\
Qwen3.5 4B
& 0.5000 & 0.1348 & \textbf{0.8652} & 0.0000
& 0.5000 & 0.1348 & \textbf{0.8652} & 0.0000 \\
Gemma4 E4B
& \underline{0.6835} & \textbf{0.2728} & 0.8630 & 0.3151
& \textbf{0.6908} & \textbf{0.2920} & 0.8509 & \underline{0.3300} \\
Gemma4 E2B
& 0.4796 & 0.1320 & 0.6689 & 0.2973
& 0.6432 & 0.2074 & 0.8191 & 0.3207 \\
MedGemma
& 0.5002 & 0.1349 & \underline{0.8648} & 0.0000
& 0.5002 & 0.1349 & \underline{0.8648} & 0.0000 \\

\midrule
\multicolumn{9}{l}{\textbf{\texttt{\texttt{CTPA+EHR}}}} \\
Qwen3.5 4B
& 0.5000 & 0.1348 & \textbf{0.8652} & 0.0000
& 0.5000 & 0.1348 & \textbf{0.8652} & 0.0000 \\
Gemma4 E4B
& \textbf{0.6952} & \underline{0.2609} & 0.8464 & \underline{0.3234}
& \underline{0.6886} & \underline{0.2776} & 0.8416 & \textbf{0.3502} \\
Gemma4 E2B
& 0.6261 & 0.1866 & 0.8097 & \textbf{0.3298}
& 0.6432 & 0.2074 & 0.8191 & 0.3207 \\
MedGemma
& 0.5000 & 0.1348 & \textbf{0.8652} & 0.0000
& 0.5000 & 0.1348 & \textbf{0.8652} & 0.0000 \\

\bottomrule
\end{tabular}
\end{adjustbox}
\end{table}

\subsection{Aggregate Performance Across Modalities and Prompting Settings}
\label{sec:aggregate_results}

Table~\ref{tab:main_results_zero_fewshot} summarizes the aggregate test performance across all eight INSPECT tasks. Overall, compact multimodal models differ substantially in their ability to use the available clinical evidence. Qwen3.5 4B remains close to chance level discrimination across modality and prompting settings, with minimal recovery of positive cases. MedGemma shows a similar pattern, with little evidence of improvement after adding EHR evidence or four-shot demonstrations. These results suggest that the models are not simply failing to follow the output format. Rather, they often settle into majority class behavior that appears superficially accurate under imbalanced labels but fails to identify positive cases.

The Gemma models show a different pattern. In the zero-shot setting, Gemma4 E4B improves substantially when CTPA and EHR evidence are combined, reaching the best aggregate zero-shot AUROC of 0.6893 and F1 of 0.3286 under \texttt{\texttt{CTPA+EHR}} input. Gemma4 E2B also benefits from additional clinical context, moving from chance level discrimination in \texttt{CTPA-Only} prompting to stronger performance when EHR evidence is available. These trends suggest that the aggregate signal in INSPECT is accessible to compact models, but that successful use of this signal depends strongly on model family and input representation.

Four-shot prompting provides a selective benefit. It does not meaningfully improve Qwen3.5 4B or MedGemma, both of which remain close to majority class behavior. In contrast, it improves the strongest Gemma configurations, especially when EHR evidence is available. Gemma4 E4B with \texttt{\texttt{CTPA+EHR}} reaches the best aggregate four-shot F1 of 0.3359, while Gemma4 E2B with \texttt{\texttt{CTPA+EHR}} achieves the best aggregate four-shot AUPRC of 0.2706. These gains suggest that in-context examples help most when the model already has a usable representation of the clinical evidence. Four-shot prompting appears to refine task alignment and recovery of positive cases, rather than creating clinical reasoning ability where the underlying model has none.

The aggregate results also clarify the limited value of \texttt{CTPA-Only} prompting in this benchmark. Across models, \texttt{CTPA-Only} settings remain close to random discrimination, even when four-shot demonstrations are added. This may reflect limitations in compact visual encoders, the difficulty of representing volumetric CTPA information in a compact input interface, or the fact that several INSPECT endpoints require longitudinal clinical context beyond the image. EHR-Only and \texttt{\texttt{CTPA+EHR}} settings are consistently stronger for Gemma models, indicating that structured clinical history carries essential information for both diagnosis and prognosis.

\subsection{Diagnosis Is Substantially Easier Than Longitudinal Prognosis}
\label{sec:diagnosis_results}

Table~\ref{tab:diagnosis_results} reports task level performance for current PE diagnosis. Compared with the aggregate benchmark, this task shows much stronger performance for Gemma models. Gemma4 E4B achieves 0.9683 AUROC and 0.9216 F1 under zero-shot \texttt{\texttt{CTPA+EHR}} input, and 0.9706 AUROC under four-shot \texttt{\texttt{CTPA+EHR}} input. EHR-Only prompting is also highly competitive, with Gemma4 E4B reaching 0.9681 AUROC and 0.9151 F1 in the four-shot setting. Gemma4 E2B follows the same trend, improving from 0.8497 AUROC in zero-shot \texttt{\texttt{CTPA+EHR}} to 0.9467 AUROC in four-shot \texttt{\texttt{CTPA+EHR}}.

These results suggest that PE diagnosis is the most accessible task for compact models when EHR evidence or combined \texttt{\texttt{CTPA+EHR}} evidence is available. The high performance of EHR-Only and \texttt{\texttt{CTPA+EHR}} settings implies that the serialized clinical context contains strong diagnostic signals, and that Gemma models can map these signals to the binary diagnostic endpoint. At the same time, \texttt{CTPA-Only} performance remains weak and unstable, with limited discrimination for Gemma4 E4B and only modest discrimination for Gemma4 E2B in zero-shot inference. This contrast indicates that the benchmark should not be interpreted as a pure image recognition task. The strongest diagnostic performance arises when compact models can exploit clinically structured evidence rather than relying on visual input alone.

\subsection{Mortality Prediction Benefits From Clinical Context}
\label{sec:mortality_results}

Table~\ref{tab:mortality_results} reports performance on mortality prediction, averaged over 1 month, 6 month, and 12 month in hospital mortality tasks. Mortality prediction is more difficult than current PE diagnosis, but the strongest Gemma configurations still show meaningful discrimination. Under \texttt{\texttt{CTPA+EHR}} input, Gemma4 E4B achieves the best zero-shot mortality AUROC of 0.7242 and the best four-shot mortality AUROC of 0.7240. Its F1 remains the highest in both settings, reaching 0.3049 in zero-shot and 0.2966 in four-shot evaluation. Gemma4 E2B also improves markedly when clinical context is available, rising from weak \texttt{CTPA-Only} performance to 0.6789 AUROC under four-shot \texttt{\texttt{CTPA+EHR}} input.

The mortality results further emphasize the importance of longitudinal patient information. Across most \texttt{CTPA-Only} settings, models remain close to chance level discrimination, suggesting that imaging evidence alone is insufficient for reliable mortality risk estimation in compact multimodal models. In contrast, EHR-Only prompting improves Gemma4 E4B to 0.6855 AUROC in the zero-shot setting and 0.6995 AUROC in the four-shot setting, while \texttt{\texttt{CTPA+EHR}} yields the strongest overall mortality performance. This pattern is clinically plausible: mortality after suspected PE is shaped not only by embolic findings, but also by baseline health status, prior instability, and comorbidity burden. The results therefore support the value of longitudinal and multimodal clinical evidence, while also showing that compact models require appropriate modality alignment to use this evidence effectively.

\subsection{Readmission Remains the Most Difficult Task Group}
\label{sec:readmission_results}

Table~\ref{tab:readmission_results} shows that readmission prediction is the hardest task group in our benchmark. Even the strongest configurations produce substantially lower AUROC and F1 than those observed for PE diagnosis or mortality prediction. In zero-shot evaluation, Gemma4 E4B with \texttt{\texttt{CTPA+EHR}} achieves the best readmission AUROC of 0.5593, while four-shot \texttt{\texttt{CTPA+EHR}} prompting with Gemma4 E2B reaches the best AUROC of 0.6141 and the best F1 of 0.2221. Although this represents an improvement over chance level baselines, the absolute performance remains modest.

This result shows that the benchmark is not simply measuring whether a model can detect obvious PE related cues. Readmission is a complex downstream outcome shaped by social, clinical, institutional, and longitudinal factors that may not be fully captured by a short serialized patient history or a single imaging event. The gap between diagnosis and readmission also explains why macro averaged results are much lower than the best task specific diagnostic numbers. A model can perform strongly on PE diagnosis while still struggling with longer horizon utilization outcomes. This task level heterogeneity motivates reporting per group results rather than relying only on aggregate averages.

\subsection{Pulmonary Hypertension Shows Intermediate Difficulty}
\label{sec:ph_results}

Table~\ref{tab:ph_results} reports performance on 12 month pulmonary hypertension prediction. This endpoint is more difficult than current PE diagnosis but generally more tractable than readmission. Gemma4 E4B achieves the strongest zero-shot AUROC of 0.6952 under \texttt{\texttt{CTPA+EHR}} input and the strongest four-shot AUROC of 0.6908 under EHR-Only input. The best F1 appears under Gemma4 E4B with four-shot \texttt{\texttt{CTPA+EHR}} input, reaching 0.3502. Gemma4 E2B also performs competitively in \texttt{\texttt{CTPA+EHR}} and EHR-Only settings, although it remains below Gemma4 E4B in discrimination.

The pulmonary hypertension results suggest that longer horizon cardiopulmonary risk can be partially captured by compact models when clinically relevant context is available. Unlike readmission, pulmonary hypertension is more directly related to cardiopulmonary disease state and may therefore be more predictable from the combination of imaging findings and longitudinal EHR signals. However, \texttt{CTPA-Only} settings again show weak discrimination, reinforcing the conclusion that visual input alone is not sufficient for compact model prognosis in this benchmark.

\subsection{Failure Modes and Modality Effects}
\label{sec:failure_modes}

Across all tables, the dominant failure mode is majority class collapse. Qwen3.5 4B and MedGemma often achieve high accuracy while remaining close to chance level discrimination with minimal recovery of positive cases. This is most visible in the aggregate results in Table~\ref{tab:main_results_zero_fewshot}, but it also appears repeatedly in the task group tables. Because many INSPECT endpoints are imbalanced, a model can obtain high accuracy by predicting the negative class for most cases. Such behavior is clinically unhelpful because the positive class corresponds to the patients most relevant for diagnosis, risk assessment, or follow up.

The modality comparison reveals a second failure mode: access to more evidence does not guarantee evidence use. \texttt{\texttt{CTPA+EHR}} improves Gemma models in most aggregate and prognostic settings, but it does not improve Qwen3.5 4B or MedGemma. Likewise, \texttt{CTPA-Only} prompting is generally weak even when four examples are provided. These results suggest that compact multimodal clinical reasoning depends on the interaction between model architecture, input representation, and task type. Multimodal inputs are useful only when the model can align the correct evidence source with the clinical endpoint.

Finally, few-shot prompting should be interpreted as a limited adaptation mechanism rather than a universal solution. In several Gemma settings, four-shot prompting improves AUPRC or F1, especially for EHR and \texttt{\texttt{CTPA+EHR}} inputs. However, it does not consistently improve AUROC, and it does not help models that already show majority class behavior in zero-shot evaluation. This pattern suggests that in-context examples mainly improve output calibration and task alignment when the model has already learned a relevant representation of the input. They cannot reliably compensate for weak visual grounding, poor clinical evidence use, or majority class bias.

\subsection{Threats to Validity}
\label{sec:threats}

Several limitations should be considered when interpreting these results. First, the benchmark is evaluated on the INSPECT test split from a single health system, so performance may reflect local imaging protocols, coding practices, patient mix, and follow-up patterns. External validation is necessary before drawing broader conclusions about clinical generalization. Second, some labels in INSPECT are generated through report based or structured phenotyping pipelines rather than exhaustive manual chart review, which may introduce label noise. Third, our \texttt{CTPA-Only} setting may underestimate the potential of specialized volumetric imaging models, because compact multimodal language models are not necessarily optimized for three dimensional radiology input.

Fourth, the evaluation focuses on point estimates in the main tables; future work should report patient-level confidence intervals or paired significance tests for primary comparisons. Fifth, the EHR serialization strategy necessarily compresses long clinical histories into a bounded context window. Different serialization, truncation, or temporal summarization choices may affect performance. Finally, the present analysis evaluates structured answer accuracy and confidence based discrimination, but does not yet fully assess clinical explanation quality, calibration, robustness to prompt variation, or deployment efficiency. These dimensions remain important for future work before compact multimodal models can be considered clinically reliable. We also do not include clinician QA baselines or parameter efficient fine-tuning baselines, both of which are important for future validation beyond this exploratory benchmark.

\section{Conclusion}
\label{sec:conclusion}

We present a compact-model benchmark for PE diagnosis and prognosis using INSPECT. By reformulating eight diagnostic and prognostic endpoints as structured clinical question answering tasks, we evaluate whether small multimodal language models can use CTPA, EHR, and combined CTPA+EHR evidence under zero-shot and four-shot prompting. The results show that compact multimodal models remain fragile in this setting. Several model and modality combinations collapse toward majority-class behavior, producing high apparent accuracy but near-random AUROC and near-zero F1. At the same time, Gemma4 E4B and Gemma4 E2B demonstrate that compact models can extract meaningful clinical signal when the input modality and model capability are better aligned. The strongest completed zero-shot setting is Gemma4 E4B with \texttt{CTPA+EHR} input. In the four-shot setting, \texttt{CTPA+EHR} remains strongest overall, with Gemma4 E4B achieving the highest AUROC and F1 and Gemma4 E2B achieving the highest AUPRC.




\bibliographystyle{icml2026}
\bibliography{refs}

\end{document}